\documentclass[letterpaper, 10 pt, conference]{ieeeconf} 
\IEEEoverridecommandlockouts     
\usepackage{cite}
\usepackage{hyperref}
\usepackage[pdftex]{graphicx}
\usepackage{epstopdf}
\usepackage{amsmath}
\usepackage{amssymb}
\usepackage{array}
\usepackage{wasysym}
\usepackage{bm}
\usepackage{units}
\usepackage[per-mode=symbol]{siunitx}
\usepackage{url}
\usepackage{color}
\usepackage{epstopdf}
\usepackage[ruled]{algorithm2e}
\usepackage{multirow}
\usepackage{booktabs}
\usepackage{lineno}

\title{\LARGE \bf Computationally-Robust and Efficient Prioritized Whole-Body Controller with Contact Constraints}

\author{D. Kim, J. Lee, J. Ahn, O. Campbell, H. Hwang, and L. Sentis$^{1}$
\thanks{$^{1}$ Luis Sentis is a Faculty of Aerospace Engineering, University of Texas at Austin, TX, USA
        {\tt\small lsentis@austin.utexas.edu}}%
}

\begin{document}

\maketitle
\thispagestyle{empty}
\pagestyle{empty}

\begin{abstract}
In this paper, we devise methods for the multi-objective control of humanoid robots, a.k.a. prioritized whole-body controllers, that achieve efficiency and robustness in the algorithmic computations. We use a form of whole-body controllers that is very general via incorporating centroidal momentum dynamics, operational task priorities, contact reaction forces, and internal force constraints. First, we achieve efficiency by solving a quadratic program that only involves the floating base dynamics and the reaction forces. Second, we achieve computational robustness by relaxing task accelerations such that they comply with friction cone constraints. Finally, we incorporate methods for smooth contact transitions to enhance the control of dynamic locomotion behaviors. The proposed methods are demonstrated both in simulation and in real experiments using a passive-ankle bipedal robot.     
\end{abstract}

\section{Introduction}
During the DARPA Robotics Challenge in 2013 and 2015, various humanoid robots participating in the competition attempted to execute a variety of tasks using a class of controllers called whole-body controllers (WBC), which are easily reconfigurable to accomplish multiple control objectives under environmental constraints \cite{iagnemma2015special}. All WBCs for humanoid control share common features: accounting for system dynamics, addressing reaction forces to control the floating base, and using a mapping between the operational spaces and the configuration space. WBCs have become a standard method to control a multi-limb system executing both manipulation and locomotion. There are various types of WBCs, but for convenience, we categorize them as follows: 1) quadratic programming (QP) based WBCs without task hierarchy \cite{kuindersma2014efficiently, stephens2010dynamic, feng2015optimization, Koolen:2016ci}, 2) null space projection based WBCs \cite{sentis:2007thesis, sentis2010compliant,Lee:2015gx}, 3) hierarchical quadratic programming (HQP) \cite{escande2014hierarchical,righetti2012quadratic,Herzog:2016ce}, and 4) integrated quadratic programming and null space projection methods\cite{Henze:2016ct, Righetti:uc}.

A popular way to implement WBCs is using QPs to find the reaction forces and torque commands satisfying equality and inequality constraints. These methods have been widely adopted for real-hardware experiments. Using these algorithms on a compliant humanoid robot, \cite{stephens2010dynamic} demonstrated the ability to lift a heavy object, balance under external disturbances, and walk. With the integration of a foot stepping planner, \cite{Kuindersma:2015cw} showed robust stepping over cinder blocks and accomplished fast exploration of the rough terrain. \cite{Koolen:2016ci} showed stable locomotion on cinder blocks and rubble. By integrating the WBC method and advanced contact area estimation, \cite{Wiedebach:2016eu} accomplished walking using only partial footholds. In recent times, WBCs have focused on leveraging early capabilities like task hierarchies, and increasing computational robustness and efficiency. Our methods shown here incorporate the latest features of WBCs while achieving higher computational efficiency than state-of-the-art methods.

Among the various capabilities of WBCs, a task hierarchy is important because it provides a guarantee that higher priority tasks will not fail due to conflicts with lower priority tasks. Hierarchies are similar to a resource allocation problem in which some resources (i.e. robot appendages or parts) are more important than others while attempting to accomplish multiple goals. Null space projection based methods accomplish a strict task hierarchy by projecting the lower priority task into the null space of higher priority tasks. This method was discussed early on by \cite{siciliano1991general} and was extended to whole-body dynamic control in \cite{sentis:2007thesis}. One drawback of projection-based methods is the inability to incorporate inequality constraints. This limitation becomes a concern when robot motions are highly dynamic causing the robot to break contact. Null space based methods assume that contact constraints are equalities which is physically incorrect. 

HQP-based WBCs have been devised to enforce both task hierarchies and inequality constraints \cite{Kanoun:2011ey, deLasa:2010ek, escande2014hierarchical, Herzog:2016ce}. 
\cite{Wensing:2013fm} proposed prioritized task-space control by iteratively solving optimizations.
\cite{escande2014hierarchical} proposed an enhanced HQP by introducing a new QP formulation and a new solver that accomplishes WBC via a single QP problem instead of having multiple QPs running iteratively. \cite{Herzog:2016ce} proposed efficient ways to solve iterative QPs and showed experimental results with a torque controlled robot. HQP-based WBC is a generic formulation that can incorporate many types of constraints and tasks. However, the computational cost is high for real-time applications (although several experimental validations have achieved about 1 \si{\milli\second} computational times in full humanoid robots).

After previous studies \cite{Kim:2016tro, Zhao:2014jw}, we have learned that the feedback control performance heavily depends on control loop times. Bandwidth differences are more noticeable when the motion is highly dynamic. For example, in the demonstration of point-foot biped stabilization \cite{Kim:2016tro} using WBCs, increasing the update rate from 1\si{\kilo\hertz} to 1.5\si{\kilo\hertz} provided noticeably larger posture and foot position control bandwidth. Moreover, fast WBCs are also desirable when considering other problems that require heavy computations, such as model-predictive control, trajectory optimizations, and machine learning for motion generation. In this paper, we integrate QP-based methods and null space projection WBC methods within a formulation that achieves higher computational efficiency than leading WBC methods. 

The approach presented in \cite{Henze:2016ct} is similar to ours, but we decrease the computational cost by removing operational space computations and instead computing acceleration in the configuration space based on the desired task commands. Also, the QP formulation in \cite{Henze:2016ct} can actually become infeasible due to torque limits and constraints, and their proposed method does not guarantee a strict task hierarchy. In contrast, our algorithm relaxes commanded task accelerations to make sure that the solution is always feasible. Furthermore, the torque limits in \cite{Henze:2016ct} can be violated because the torque commands for posture control are separated from the optimization, and added to the optimization result afterward. 

For computational efficiency, we simplify the formulation even further over the previous acceleration-based methods by removing the terms representing the time derivative of the prioritized task Jacobians. Besides our efforts to increase computational efficiency, our formulation also addresses internal constraints and contact force transitions.  By adding internal constraints, the torque command complies with constraints that internally exist such as mechanically coupled joints or closed kinematic chains. Our contact force transition method smooths out the jerk induced by a contact constraint change. 

In summary, our contributions are: 1) a new WBC formulation, dubbed whole-body dynamic control (WBDC), providing benefits in efficient computation and algorithmic robustness, 2) the addition of new functionalities (internal constraints, contact transitions), and 3) the validation of the proposed algorithm both in physics-based simulations and in biped robot experiments.
   
\section{Whole-Body Dynamic Control Formulation}
\begin{figure}
\centering
\includegraphics[width=1.0\columnwidth]{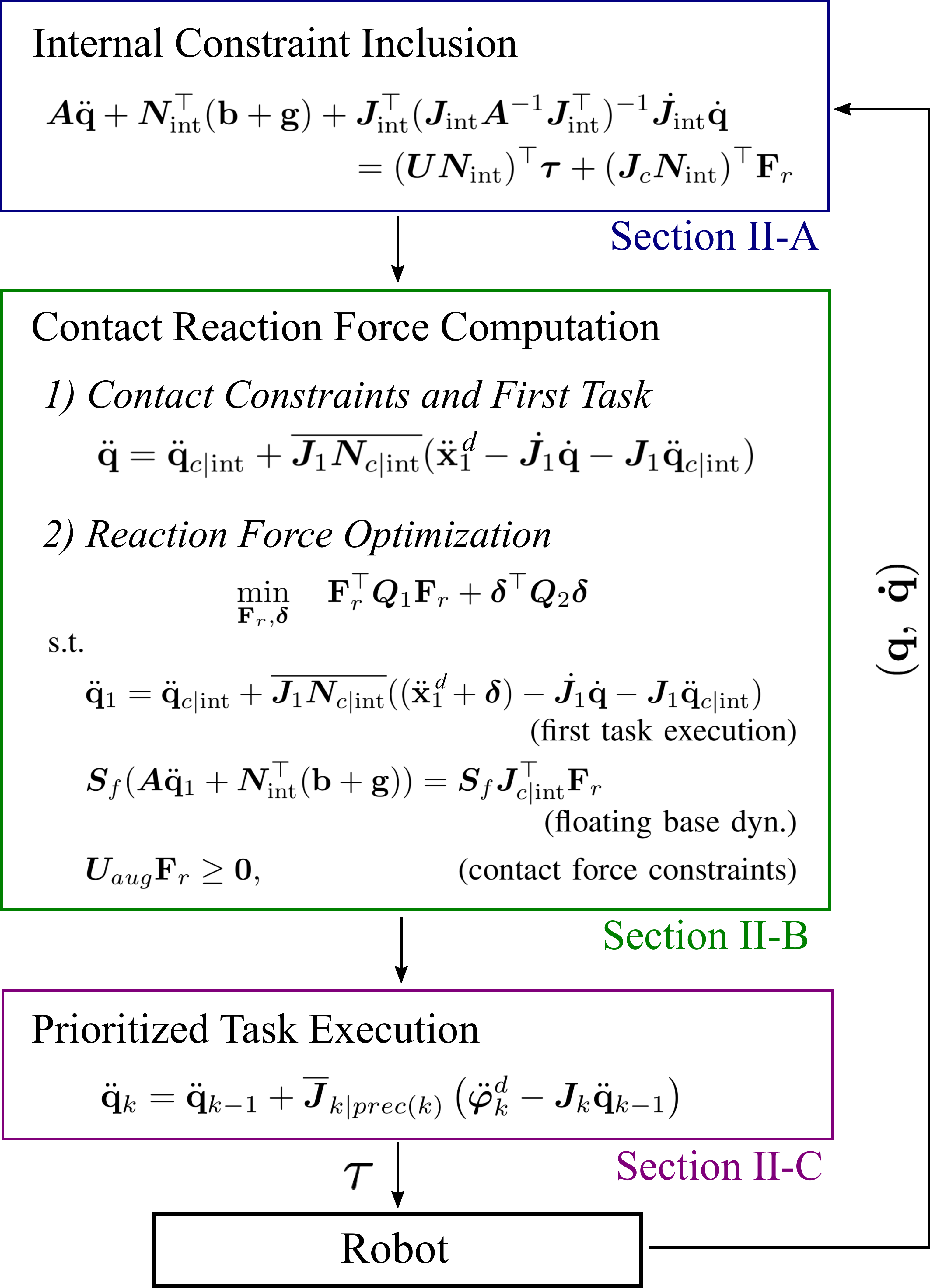}
\caption{{\bf Whole-body Dynamic Control process flow.} During every control loop, the dynamic and kinematic model of a robotic system is updated with the sensed configuration ($\mathbf{q}$, $\dot{\mathbf{q}}$). WBDC projects dynamics into the null space of internal constraints (Section.~\ref{sec:wbdc_internal}). After sequential projections addressing contact constraints and the first task are computed, a QP solver computes reaction forces and relaxed floating base task commands (Section.~\ref{sec:wbdc_qp}). WBDC goes on to compute full configuration space accelerations by using a null space projection strategy. (Section.~\ref{sec:wbdc_task_hierarchy}).}
\label{fig:wbdc}
\end{figure}

We present an algorithm for whole-body control, dubbed WBDC, that improve over the state-of-the-art on efficiency and computational robustness. Fig.~\ref{fig:wbdc} illustrates the overall process performed by WBDC. The details are explained below.

\subsection{Internal Constraints Inclusion}
\label{sec:wbdc_internal}
We prescribe constraints into two categories: 1) internal and 2) contact constraints. Internal constraints are holonomic and do not influence the floating base dynamics. Contact constraints involve reaction forces from the environment, and thus, exert forces on the floating base. Examples of internal constraints include bilateral constraints in coupled joints (e.g., hip yaw joints in the NAO \cite{strom2009omnidirectional} and waist joints in the Dreamer robot \cite{sentis2013implementation}) and closed chains in delta robots \cite{pierrot1990delta}.  

The first step in WBDC is projecting the multi-body dynamics into the null space of the internal constraints to remove the term representing the internal forces incurred by the constraint.
\begin{equation}
\bm{A} \ddot{\mathbf{q}} + \mathbf{b}+ \mathbf{g} = \bm{U}^{\top}\bm{\tau} + \bm{J}_{\rm int}^{\top}\mathbf{F}_{\rm int} + \bm{J}_c^{\top}\mathbf{F}_r,
\end{equation}
where $\bm{A}$, $\mathbf{b}$, and $\mathbf{g}$ represent mass, coriolis/centrifugal, and gravitational forces, respectively. $\bm{U}$ is a selection matrix to map actuated joint torques to the full configuration effort. $\bm{J}_{\rm int}$ and $\bm{J}_c$ are internal and contact constraint Jacobians satisfying the conditions
\begin{equation}
\label{eq:jac_constr}
\begin{split}
& \mathbf{0} = \bm{J}_{x}\dot{\mathbf{q}}, \\
& \mathbf{0} = \bm{J}_{x}\ddot{\mathbf{q}} + \dot{\bm{J}}_{x}\dot{\mathbf{q}},
\end{split}
\end{equation}
where $\bm{J}_{x}$ represents either $\bm{J}_{\rm int}$ or $\bm{J}_c$. Before presenting the projected dynamics, we show two important terms. One is a dynamically consistent inverse,
\begin{equation}
	\overline{\bm{J}} = \bm{A}^{-1} \bm{J}^{\top} \left( \bm{J} \bm{A}^{-1}\bm{J}^{\top}  \right)^{-1},
\end{equation}
and the other one is a null space projection matrix,
\begin{equation}
\bm{N} = \bm{I} - \overline{\bm{J}}\bm{J}.
\end{equation}
By using standard null space projection techniques, we obtain the following equation,
\begin{equation}
\label{eq:projected_dyn}
\begin{split}
\bm{A}\ddot{\mathbf{q}} + \bm{N}_{{\rm int}}^{\top}(\mathbf{b} + \mathbf{g})& + \bm{J}_{\rm int}^{\top} (\bm{J}_{\rm int} \bm{A}^{-1} \bm{J}_{\rm int}^{\top})^{-1} \dot{\bm{J}}_{\rm int}\dot{\mathbf{q}} \\ 
& \quad  =(\bm{U}\bm{N}_{\rm int})^{\top} \bm{\tau} + (\bm{J}_c \bm{N}_{\rm int})^{\top}\mathbf{F}_r,
\end{split}
\end{equation}
where $\bm{N}_{\rm int}$ is the null space projection matrix derived from the internal constraints.
Again, an important property of the internal constraints is that by definition they do not influence the floating base dynamics. Thus, the rows representing the floating base in $(\bm{U}\bm{N}_{\rm int})^{\top}$ are all zero. Using this property, $\bm{\tau}$ does not appear in the equality constraint of our QP formulation described in the next section. 

\subsection{Contact Reaction Forces Computation}
\label{sec:wbdc_qp}
Before finding reaction forces with QP, we need to address contact constraints. After computing the accelerations in the configuration space satisfying the contact constraints, the first task command is projected through the null space of the internal and contact constraints and added to the configuration space accelerations. Then, QP computes reaction forces satisfying unilateral and friction cone constraints. In the QP formulation, we add an upper bound on the reaction forces in the vertical direction to accomplish smooth contact force transitions.

\subsubsection{Contact Constraints and First Task}
\label{sec:wbdc_contact}
Similar to the internal constraints, contact constraints are also characterized by zero acceleration at the contact points. Therefore, all task commands need to be in the null space of the contact constraints so that resultant motions do not incur acceleration at contact points.
This is achieved by inverting the projected contact Jacobian, $\overline{\bm{J}}_{c|{\rm int}} (= \overline{\bm{J}_c\bm{N}_{\rm int}})$, and from the condition in Eq. \ref{eq:jac_constr}, we obtain
\begin{equation}
\ddot{\mathbf{q}}_{c|{\rm int}} = \overline{\bm{J}}_{c|{\rm int}}(-\dot{\bm{J}}_c\dot{\mathbf{q}}).
\end{equation}
Then, we add the desired accelerations for the first task projected through the null space of the constraints,
\begin{equation}
\label{eq:qp_prep_dyn}
\ddot{\mathbf{q}}_1 = \ddot{\mathbf{q}}_{c|{\rm int}} + \overline{\bm{J}_1\bm{N}_{c|{\rm int}}}(\ddot{\mathbf{x}}_1^d - \dot{\bm{J}}_1\dot{\mathbf{q}} - \bm{J}_1\ddot{\mathbf{q}}_{c|{\rm int}}),
\end{equation}
where $\bm{N}_{c|{\rm int}} (= \bm{I}-\overline{\bm{J}}_{c|{\rm int}} \bm{J}_{c|{\rm int}})$ is a projection matrix of constraints and $\ddot{\mathbf{x}}_1$ is the acceleration command for the first task. $\dot{\bm{J}}_1\dot{\mathbf{q}}$ and $\bm{J}_1\ddot{\mathbf{q}}_{c|{\rm int}}$ represent the influence on the first task acceleration caused by the configuration velocity and the computed acceleration incorporating constraints, respectively. Eq.~\eqref{eq:qp_prep_dyn} is similar to the equations in \cite{Righetti:uc}, but \cite{Righetti:uc} omitted the term representing the influence of the higher priority task command to the lower priority, which is $\bm{J}_1\ddot{\mathbf{q}}_{c|{\rm int}}$ in Eq.~\eqref{eq:qp_prep_dyn}.

Note that the first task must span centroidal dynamics, which is equivalent to spanning the floating base dynamics. Thus, in our simulations and experiments, we assign the first task to be either full joint position control, centroidal momentum control, or CoM and floating base orientation control task. To verify the first task spans the floating base dynamics, we check that the rank of the projected mass matrix is zero,
\begin{equation}
{\rm rank}(\bm{S}_f\bm{A}\bm{N}_{1|c|{\rm int}}) = 0,
\end{equation}
where $\bm{S}_f$ is a selection matrix to select the configuration representing the floating base. In general cases, the matrix selects the first six rows of $\bm{A}$. $\bm{N}_{1|c|{\rm int}}$ is the null space projection matrix of the projected first task. 

\subsubsection{Reaction Force Computation}
\label{sec:wbdc_optimization}
After obtaining the joint acceleration accounting for the constraints and the first task, we are now ready to compute the reaction forces. We use an open-source QP solver and follow the method proposed in \cite{Goldfarb:1983ik}. To ensure that the QP problem is feasible, we relax the first task command. The formulation of our QP problem is
\begin{equation} \label{eq:qp_cost}
\min_{\mathbf{F}_r, \bm{\delta}}\quad \mathbf{F}_{r}^{\top} \bm{Q}_1 \mathbf{F}_{r} + \bm{\delta}^{\top}\bm{Q}_2\bm{\delta}\vspace{0.7mm} \\
 \vspace{-4mm}
\end{equation}
\begin{align*}
\text{s.t.} & \\
\tag{first task execution}
&\ddot{\mathbf{q}}_1 = \ddot{\mathbf{q}}_{c|{\rm int}} + \overline{\bm{J}_1 \bm{N}_{c|{\rm int}}}
\left( 
(\ddot{\mathbf{x}}_1^d + \bm{\delta}) - \dot{\bm{J}}_1\dot{\mathbf{q}} - \bm{J}_1\ddot{\mathbf{q}}_{c|{\rm int}}
\right)\\
\tag{floating base dyn.}
&\bm{S}_f 
\left(
\bm{A} \ddot{\mathbf{q}}_1 + \bm{N}_{\rm int}^{\top} (\mathbf{b} + \mathbf{g})
\right) 
= \bm{S}_f \bm{J}_{c|{\rm int}}^{\top} \mathbf{F}_r \\
\tag{contact force constraints}
 & \bm{W}_{aug} \mathbf{F}_r \geq \mathbf{0},
\end{align*}
where $\mathbf{F}_r$, $\bm{\delta}$, and $\bm{W}_{aug}$ are reaction forces, relaxation variables, and an augmented contact constraint matrix, respectively. When describing the floating base dynamics, we remove the terms $\bm{J}_{\rm int}^{\top}(\bm{J}_{\rm int}\bm{A}^{-1}\bm{J}_{\rm int}^{\top})^{-1}\dot{\bm{J}}_{\rm int}\dot{\mathbf{q}}$ and $(\bm{U}\bm{N}_{\rm int})^{\top}\bm{\tau}$ of Eq.~\eqref{eq:projected_dyn}, since internal constraints have all zero terms in the rows corresponding to the floating base dynamics: 
\begin{equation}
\begin{split}
&\bm{S}_f\bm{J}_{\rm int}^{\top}(\bm{J}_{\rm int}\bm{A}^{-1}\bm{J}_{\rm int}^{\top})^{-1}\dot{\bm{J}}_{\rm int} = \bm{0}\\ 
&\bm{S}_f(\bm{U}\bm{N}_{\rm int})^{\top}=\bm{0}.
\end{split}
\end{equation}
The contact force constraints in the QP shown in Eq. \ref{eq:qp_cost} include the friction cone inequalities to prevent slipping, and a reaction torque limit to prevent flipping or rolling the contacting link. In the case of a surface contact with a rectangular support area, the friction cone inequalities are
\begin{equation}
\label{eq:contact_constraint}
\mid f_x \mid \leq \mu f_z, \quad \mid f_y \mid \leq  \mu f_z,
\quad f_z > 0.
\end{equation}
and the no-flip condition inequalities are
\begin{align}
\label{eq:flip}
\begin{split}
\mid \tau_x \mid  \leq d_y f_z, & \quad \mid \tau_y \mid  \leq d_x f_z, \\
\tau_{z, {\rm min}} \leq &~ \tau_z \leq \tau_{z, {\rm max}},
\end{split}
\end{align}
where
\begin{equation}
\begin{split}
\tau_{z,{\rm min}} &\triangleq -\mu(d_x+d_y)f_z+\mid d_y f_x-\mu\tau_x\mid+\mid d_x f_y-\mu\tau_y\mid, \\
\tau_{z,{\rm max}} &\triangleq \mu(d_x+d_y)f_z-\mid d_y f_x+\mu\tau_x\mid-\mid d_x f_y+\mu\tau_y\mid,
\end{split}
\end{equation}
and $d_x,~d_y$ represent the distance from the center of the rectangle to the vertex in the local contact frame. Note that, $[\tau_{x},~\tau_{y},~\tau_{z},~f_x,~f_y,~f_z]^{\top}$ is the wrench representing resultant forces on the surface taken at the center of the rectangle in the local coordinate frame. \cite{Caron:2015tf} describes the closed-form formula for the contact wrench cone $\bm{W} \in \mathbb{R}^{17 \times 6}$.
Then, the contact constraints for the $i$-th contact are compactly expressed as
\begin{equation}
\bm{W}_{i} \mathbf{F}_r \geq \mathbf{0}.
\end{equation}
The final step is to include contact frame rotations and augment to multiple contact points. For example, the formulation for $k$ contact points is
\begin{equation}
\label{eq:W_aug_matrix}
\bm{W}_{aug} \mathbf{F}_r = \begin{bmatrix} \bm{W}_{1}\bf{R}_{aug,1}&0&0 \\ 0& \ddots &0 \\ 0&0&\bm{W}_{k}\bf{R}_{aug,k} \end{bmatrix} \begin{bmatrix} \mathbf{F}_{r,1} \\ \vdots \\ \mathbf{F}_{r,k}\end{bmatrix} \geq \mathbf{0},
\end{equation}
where
\begin{equation}
\bf{R}_{aug,i}=\begin{bmatrix} \bm{R}^{cm}_{i} & \bm{0}^{3 \times 3} \\ \bm{0}^{3 \times 3} & \bm{R}^{cm}_{i} \end{bmatrix}.
\end{equation}
Note that $\bm{R}^{cm}_{i}$ is SO(3) from the CoM frame to the local contact coordinate frame and allows contact constraints to be expressed locally.

When selecting the weight matrices $\bm{Q}_1$ and $\bm{Q}_2$ for the QP cost function in Eq. \ref{eq:qp_cost}, we recommend weighting $\bm{Q}_2$ more heavily than $\bm{Q}_1$. For example, if we relax the height control task, then the QP will adjust the task to allow the robot to fall below the commanded height to reduce the vertical directional reaction forces because these forces are the largest numbers among the optimization variables in general.

\subsubsection{Contact Constraint Transitions}
\label{sec:contact_transition}
WBDC also allows for optional smoothing of contact transitions by bounding the magnitude of reaction forces during the transition period. When using this functionality, we add the following inequality for the vertical directional reaction forces for each contact point:
\begin{equation}
\mathbf{F}_{r,z} \leq h\mathbf{F}_{r,z }^{\rm max} + (1-h)\mathbf{F}_{r,z}^ {\rm min},
\end{equation}
where $h \in \begin{bmatrix} 0, & 1\end{bmatrix}$. This constraint is included as an extra row in each $\bm{W}_{i}$ matrix shown in Eq \ref{eq:W_aug_matrix}. During the transition period, users increase or decrease $h$ to avoid a sudden jump in the reaction forces when the contact is engaged or disengaged. Non-vertical reaction forces are bounded by the magnitude of the reaction force in the vertical direction, thus a single additional inequality constraint for each contact point is sufficient.

At this point the QP problem is fully specified. Running a solver will return the contact reaction forces $\mathbf{F}_{r}$ and the relaxation variables $\bm{\delta}$. Plugging the latter back into the \textit{first task execution} constraint in Eq. \ref{eq:qp_cost} will yield $\ddot{\mathbf{q}}_1$.

\subsection{Prioritized Task Executions}
\label{sec:wbdc_task_hierarchy}

\begin{figure}
\begin{minipage}[t]{1.0\linewidth}
  \begin{center}
\includegraphics[width=0.85\columnwidth]{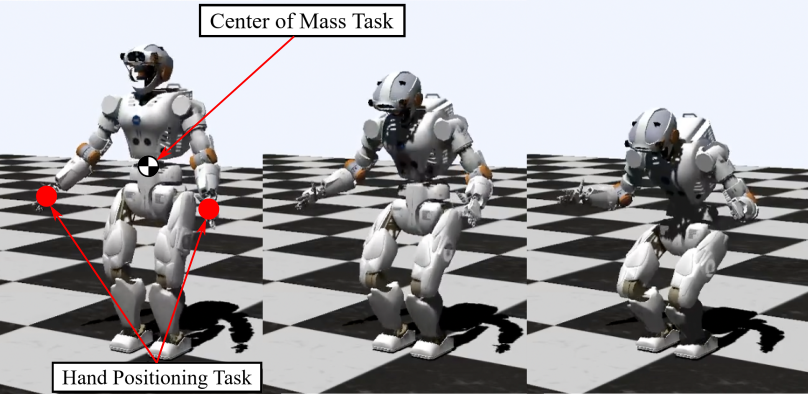}
  \end{center}
\end{minipage}
\begin{minipage}[t]{1.0\linewidth}
  \begin{center}
\includegraphics[width=1.0\columnwidth]{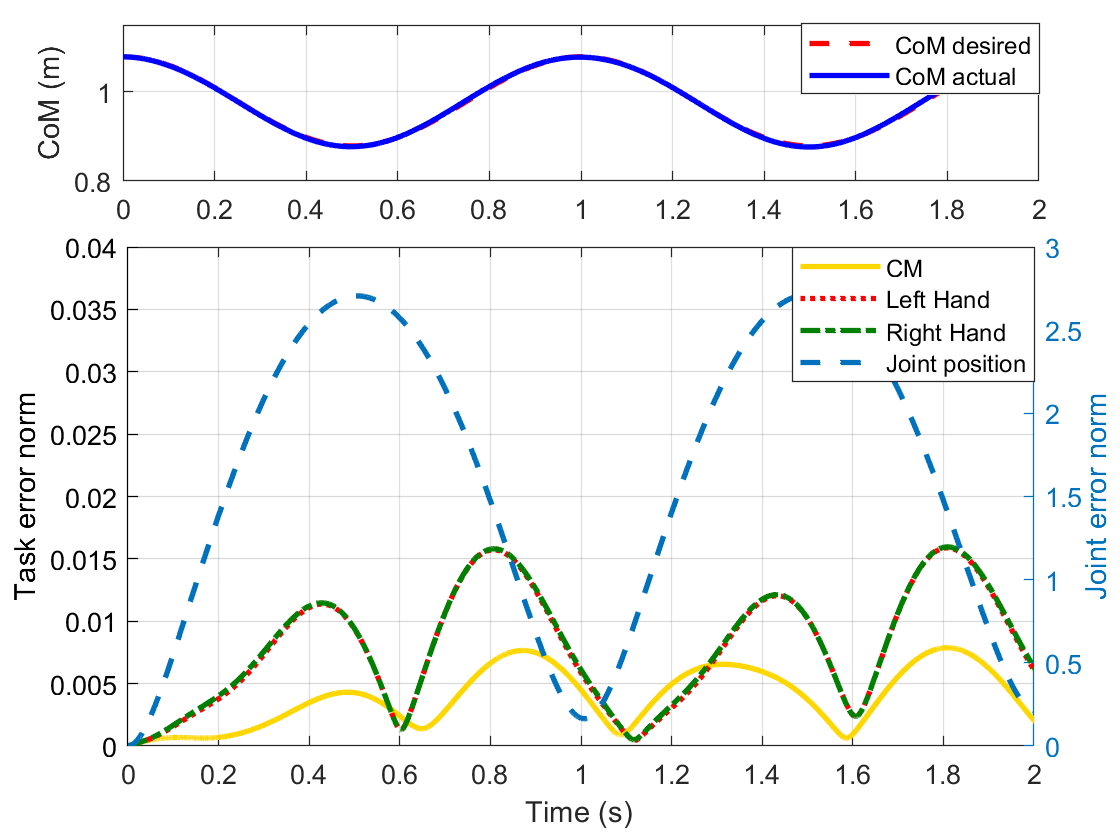}
  \end{center}
\end{minipage}
\caption{{\bf CoM height control while maintaining hand positions.} With both hand position fixed, Valkyrie moves up and down as the commanded CoM height changes. The data shows that Valkyrie's CoM height is almost identical to the desired trajectory. The norm of each task error becomes larger as the priority of the task becomes lower.}
\label{fig:valkyrie_height_ctrl}
\end{figure}

With $\ddot{\mathbf{q}}_1$, we can begin the process of iteratively projecting and applying the remaining prioritized tasks. Task-level controllers are computed in operational space as acceleration commands and converted to accelerations in configuration space using differential forward kinematics via
\begin{equation}
\begin{split}
	\dot{\mathbf{x}}_{\rm task} &= \bm{J}_{\rm task} \dot{\mathbf{q}},\\
	\ddot{\mathbf{x}}_{\rm task} &= \bm{J}_{\rm task}\ddot{\mathbf{q}} + \dot{\bm{J}}_{\rm task}\dot{\mathbf{q}}, 
\end{split}
\end{equation}
where $\mathbf{x}_{\rm task} \in \mathbb{R}^{n}$ and $\mathbf{q} \in \mathbb{R}^{m}$ represent the task's operational coordinates and the robot configuration, respectively, and $\bm{J}_{\rm task}$ is the corresponding Jacobian matrix. The joint acceleration resulting from the second desired task acceleration, $\ddot{\mathbf{x}}_{2}^{d}$ can be resolved as
\begin{align}
	&\ddot{\mathbf{q}}_2 = \ddot{\mathbf{q}}_1 + \overline{\bm{J}_{2|1}} \left( \ddot{\bm{\varphi}}_{2}^{d} -  \bm{J}_{2} \ddot{\mathbf{q}}_{1} \right), \label{eq:two_task}
\\
& \ddot{\bm{\varphi}}_{2}^{d} \triangleq \ddot{\mathbf{x}}_{2}^{d} - \dot{\bm{J}}_{2} \dot{\mathbf{q}}, \label{eq:task_cmd}
\end{align}
where $\overline{\bm{J}_{2|1}}\triangleq \overline{\left( \bm{J}_{2} \bm{N}_{1} \right)}$ represents the Jacobian associated with the second task, $\bm{J}_2$, projected into the null space of the first task, $\bm{N}_1=\bm{I} -  \overline{ \bm{J}_{1}\bm{N}_{c|{\rm int}}} (\bm{J}_{1}\bm{N}_{c|{\rm int}})$. $\bm{N}_1$ by definition is orthogonal to the Jacobian associated with the projected first task, $\bm{J}_{1}\bm{N}_{c|{\rm int}}$. Eq. (\ref{eq:two_task}) can be extended to the general $k$-th task case, using the following hierarchy
\begin{equation}
	\ddot{\mathbf{q}}_{k} = \ddot{\mathbf{q}}_{k-1} + \overline{\bm{J}}_{k|prec(k)}\left( \ddot{\bm{\varphi}}_{k}^{d} - \bm{J}_{k} \ddot{\mathbf{q}}_{k-1} \right)
\label{eq:n_tasks}
\end{equation}
with
\begin{equation}
\begin{split}
	&\bm{J}_{k|prec(k)} = \bm{J}_{k} \bm{N}_{prec(k)}, \\
    &\bm{N}_{prec(k)} = \prod_{s=2}^{k-1} \bm{N}_{s|prec(s)}\\
    &\bm{N}_{s|prec(s)} = \bm{I} - \overline{\bm{J}}_{s|prec(s)} \bm{J}_{s|prec(s)}
\end{split}
\end{equation}
where $k\geq 2$, $\bm{N}_{prec(2)} = \bm{N}_1$. This task hierarchy is similar, but not identical to \cite{siciliano1991general,Righetti:uc}. Compared to those, our method is more concise, resulting in lower computation load for similar control specifications. In particular, we do not include the computation of time derivatives of prioritized Jacobians. Previous studies use $\dot{\bm{J}}_{k|prec(k)}\dot{\mathbf{q}}$ rather than $\dot{\bm{J}}_{k}\dot{\mathbf{q}}$, which appears in Eq.~\eqref{eq:task_cmd}. We use $\dot{\bm{J}}_{k}\dot{\mathbf{q}}$ because it correctly addresses the influence of the velocity on the task space. 


\begin{figure*}
\centering
\includegraphics[width=1.95\columnwidth]{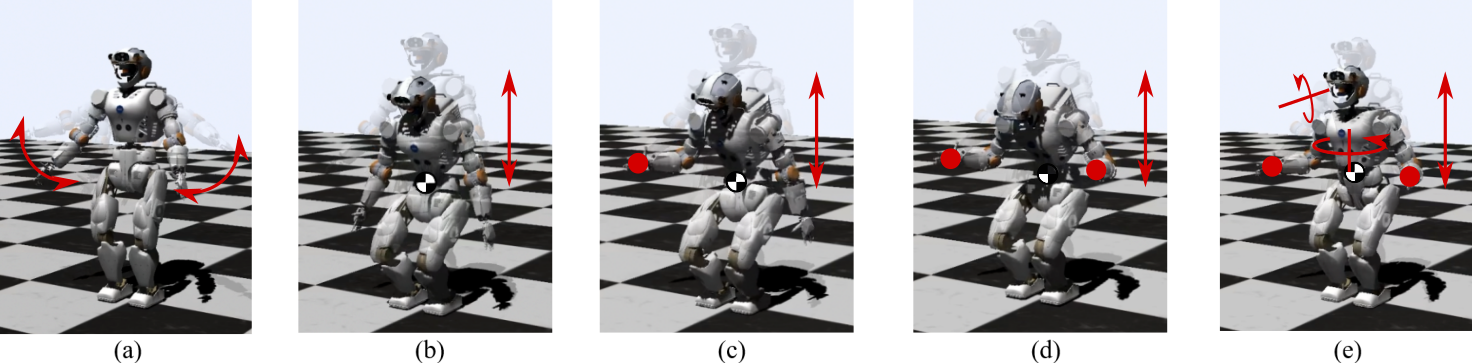}
\caption{{\bf Implementation of various task sets:} Joint position (JP), Centroidal momentum (CM), Right hand position (RHP), Left hand position (LHP), Body orientation (BO), Head orientation (HO). (a) JP (b) CM + JP (c) CM + RHP + JP (d) CM + RHP + LHP + JP (e) CM + RHP + LHP + BO + HO + JP.}
\label{fig:various_tasks}
\end{figure*}

\begin{figure}
\begin{minipage}[t]{1.0\linewidth}
  \begin{center}
\includegraphics[width=1.0\columnwidth]{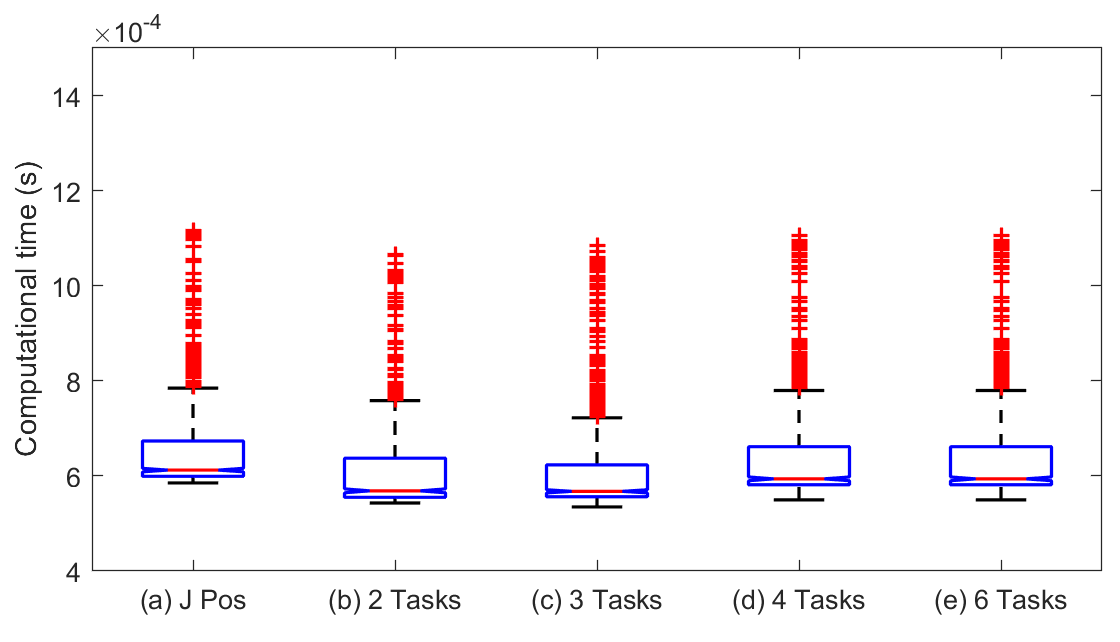}
  \end{center}
\end{minipage}
\caption{{\bf Computational time comparison:} The computation times of WBDC are plotted with respect to the task sets of the previous figure. $1000$ iterations are considered for each case set. }
\label{fig:computation}
\end{figure}

The last step of WBDC is to compute a torque command from the contact reaction forces ($\mathbf{F}_r$) and the configuration space acceleration ($\ddot{\mathbf{q}}$). By plugging these two terms into Eq.~\ref{eq:projected_dyn}, we obtain 
\begin{equation}
\begin{split}
(\bm{U}\bm{N}_{\rm int})^{\top} &\bm{\tau} =
\bm{A}\ddot{\mathbf{q}}  + \bm{N}_{\rm int}^{\top} (\mathbf{b} + \mathbf{g}) \\
& + \bm{J}_{\rm int}^{\top} (\bm{J}_{\rm int} \bm{A}^{-1} \bm{J}_{\rm int}^{\top})^{-1} \dot{\bm{J}}_{\rm int}\dot{\mathbf{q}} 
 - \bm{J}_c^{\top}\mathbf{F}_r.
\end{split}
\end{equation}
Since all terms on the right hand side are known, we can calculate the torque command by multiplying the right hand side by the inverse of $(\bm{U}\bm{N}_{\rm int})^{\top}$. To enhance the computational efficiency, we reduce the matrix by removing all zero terms corresponding to the floating base. 

\section{Simulation Results}
To verify the proposed methods, we perform physics-based simulations of Valkyrie, an adult size humanoid robot developed by NASA Johnson Space Center \cite{Radford:2015ca}. In our simulation, we reduced the number of DOFs of the robot by assuming that hand and wrist joints are fixed. Except for the reduced number of actuated joints, we built the Valkyrie simulation based on data provided by NASA. The simulations are designed to verify our claims: 1) execution of prioritized tasks, and 2) computational efficiency.

\subsection{Multiple Task Execution}

The proposed method is verified using complex whole body motions with double contact conditions as shown Fig. \ref{fig:valkyrie_height_ctrl} and Fig. \ref{fig:various_tasks}. In our simulation, Valkyrie has 28 actuated joints and the total dimensionality of the configuration space is $34$ including the $6$ virtual joints representing the floating base. The tasks used in the simulation are listed and the subscript indices indicate the priority of each task.
 
\begin{itemize}
\item[$\cdot$] $\ddot{\mathbf{x}}_{1}^d \in \mathbb{R}^{6}$: centroidal momentum (CM) \vspace{0.8mm}
\item[$\cdot$] $\ddot{\mathbf{x}}_{2}^d \in \mathbb{R}^{3}$: right hand position (RHP) \vspace{0.8mm}
\item[$\cdot$] $\ddot{\mathbf{x}}_{3}^d \in \mathbb{R}^{3}$: left hand position (LHP) \vspace{0.8mm}
\item[$\cdot$] $\ddot{\mathbf{x}}_{4}^d \in \mathbb{R}^{28}$: full joint posture (JP) \vspace{0.8mm}
\end{itemize}

In Fig.~\ref{fig:valkyrie_height_ctrl}, the CoM is commanded to move up and down while maintaining constant hand positions and double support contact. 
As explained in previous sections, the contact constraints are given the highest priority and the accelerations at the contact points are assumed to be zero. 
The first task is the CM task and is subordinate to the contact constraints. The CM task was designed to minimize the centroidal angular momentum while following a planned CoM trajectory. The vertical directional CoM trajectory is a sinusoidal function with $0.1\si{\meter}$ amplitude and $1.0\si{\hertz}$ frequency. Next, the hand position tasks are defined to maintain a constant position. The last task specification is the joint position task for all joints and the corresponding desired values are the initial joint positions. 

Fig.~\ref{fig:valkyrie_height_ctrl} shows the simulation results for the set of tasks defined above. As shown in the snapshots and the upper graph of Fig.~\ref{fig:valkyrie_height_ctrl}, the robot successfully moves up and down to track the desired trajectory while maintaining constant hand positions. Since the sum of the task dimensions is larger than the total DOF of the robot, some lower priority tasks cannot be completely executed. For this reason, the joint position task with the lowest priority has significant error ($\|\mathbf{e}\|_{2} \leq 3$), compared with errors of other tasks ($\|\mathbf{e}\|_{2} \leq 0.02$) as shown in the second graph. 

\begin{figure*}
\centering
\includegraphics[width=1.9\columnwidth]{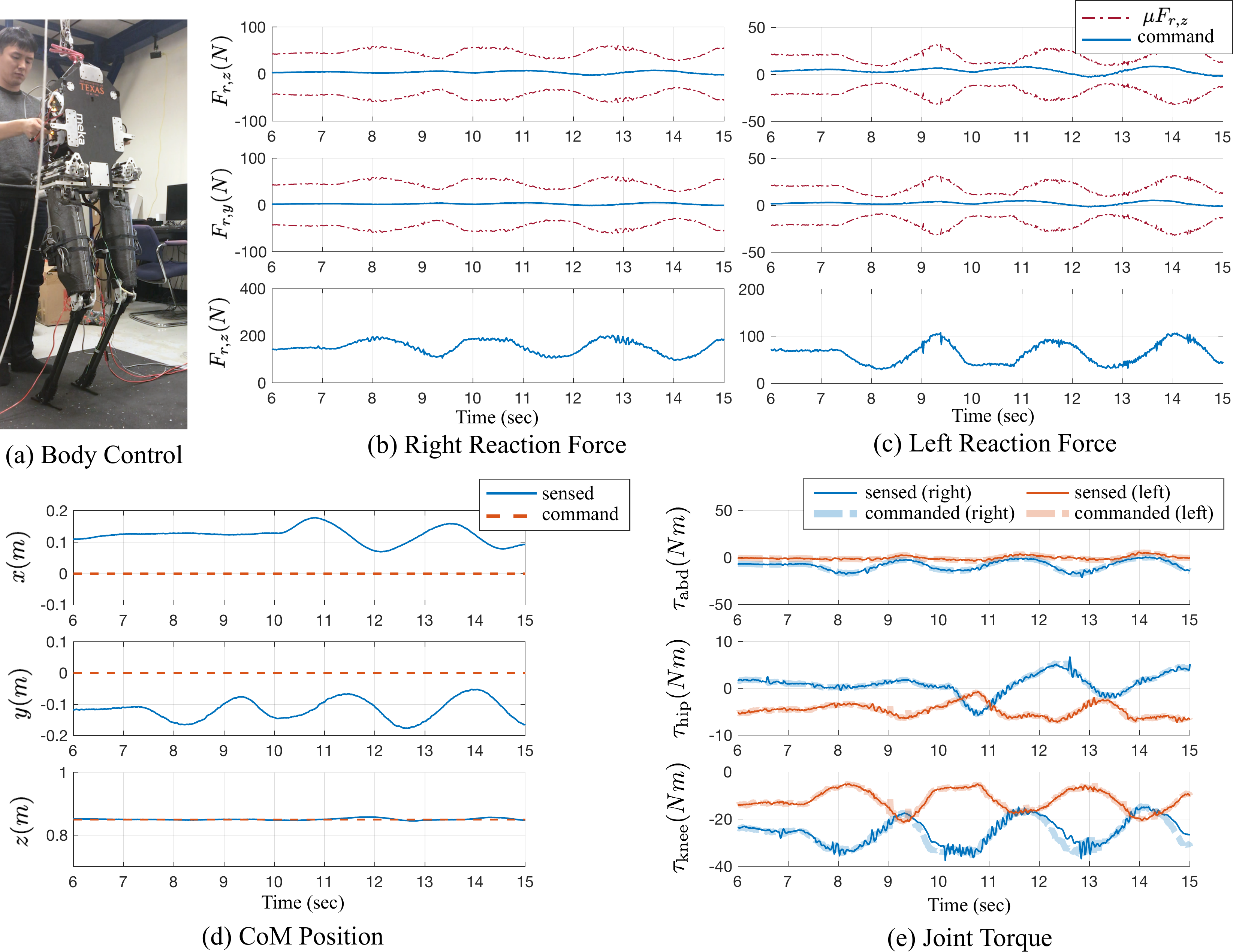}
\caption{{\bf Body Pose Control Experiment.} (a) A person grabs handles of the body and move the body to the left and to the right. (b)(c) horizontal reaction forces satisfy friction cone constraints. The friction coefficient $\mu$ is set by 0.3. (d) $x$ and $y$ directional CoM position way off from the desired position control, which is an intended behavior. (e) DOB based torque feedback controller enables high accuracy torque tracking.}
\label{fig:body_pos_ctrl}
\end{figure*}

\subsection{Computation Time Evaluation}

\begin{table}
\centering
\begin{tabular}{>{\centering}m{0.5\columnwidth}  %
                >{\centering}m{0.3\columnwidth} %
                @{}m{0pt}@{}}
\specialrule{1.5pt}{1pt}{1pt}
\vspace{1mm}
{Tasks}
& {mean / SD (\si{\milli\second})}
&\\[2.0mm] 
\hline
\hline
JPos & 0.632 / 0.0948 &\\ [1mm] \hline
CM + JPos & 0.612 / 0.0971 & \\ [1mm] \hline
CM + RHP + JP & 0.609 / 0.0999 & \\ [1mm] \hline
CM + RHP + LHP + JP & 0.636 / 0.0957 & \\ [1mm] \hline
CM + RHP + LHP + BO + HO + JP & 0.637 / 0.0957 & \\ [1mm] \hline
\vspace{-0.5mm}
\end{tabular}
\caption{Computation Time Evaluation}
\label{tb:computation_time}
\vspace{-6mm}
\end{table}

To evaluate the computational performance of the proposed algorithm, we implemented five different simulation scenarios with varying numbers of tasks as shown in Fig.~\ref{fig:various_tasks}. All of the scenarios use joint position as the lowest priority task, and each scenario successively incorporates one or two more tasks relative to the previous setup. We then timed our algorithm running on a dual-core 3.0 GHz Intel i7 processor for each of the five scenarios. Table \ref{tb:computation_time} and Fig.~\ref{fig:computation} show the measured computation time for $1000$ iterations of the algorithm for each simulated scenario. 

Overall, the average computation time was around $0.6 \si{\milli\second}$. In the proposed algorithm, the majority of the time is used to solve the QP, so the computation time would probably scale most noticeably with the size of the constraint set and the cost function (which includes the first task). As shown in the Fig. \ref{fig:computation}, case (a) when the joint position task, which has the largest task dimension, is the first task incurs slightly more computational load than that of simulation case (b) when the CM task is the first task. Furthermore, since only the first task is included in the QP where most of the time is spent, the computation time does not increase significantly with respect to the number of hierarchical tasks from (b) to (e). Therefore, this algorithm works well even with a large number of prioritized tasks. In the current setup we did not incorporate the advanced decomposition algorithms used in \cite{escande2014hierarchical, Herzog:2016ce}, so we believe that the computational efficiency could be improved by optimizing some of the matrix calculations.

\begin{figure*}
\centering
\includegraphics[width=2.0\columnwidth]{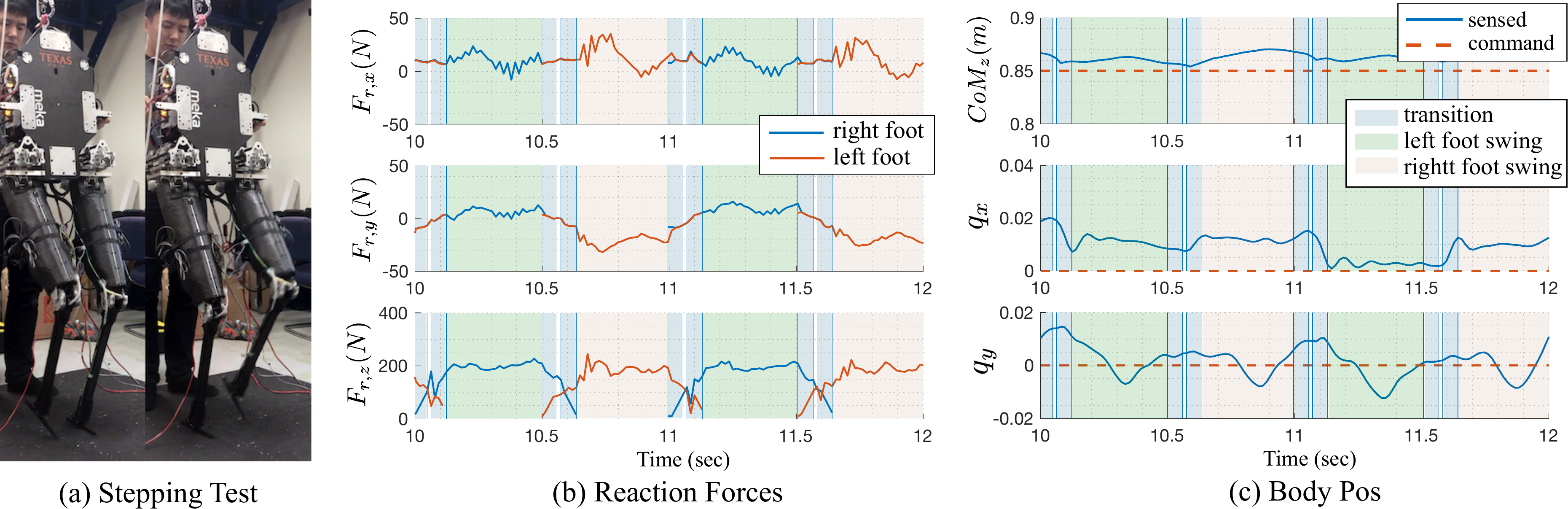}
\caption{{\bf Stepping Test Experiment.} (a) During the test, a person gently holds the robot's body to constraint horizontal movements. (b) The vertical directional reaction forces smoothly change from zero to 200\si{\newton} during transition phases, which is the period colored by blue box. (c) Mercury's CoM height, roll, and pitch are maintained around the commanded pose. $\mathbf{q}_x$ and $\mathbf{q}_y$ mean $x$ and $y$ components of quaternion.}
\label{fig:stepping_test}
\end{figure*}

\section{Passive-Ankle Biped Robot Experiments}
Our robot, Mercury, is a biped robot that is 1.5 m tall and 23 kg in weight. The leg kinematics resemble simplified human kinematics and contain, in series, an adduction/abduction hip joint, a flexion/extension hip joint, and a flexion/extension knee joint, as shown in Fig.~\ref{fig:body_pos_ctrl}(a). The lack of ankle actuation allows the carbon fiber shank to be as lightweight as possible for quick foot movements. The series elastic actuators (SEAs) on all six joints are based on a sliding spring carriage connected to the output joint by steel cables. The deformation in the springs is directly measured within the carriage assembly. For fall safety, the robot is attached to a pulley system with a block and tackle that allows us to lift the robot off of the ground.
Mercury is controlled with distributed digital signal processors connected by an EtherCAT network to a centralized PC running a RT-Preempt Linux kernel. Each DSP controls a single actuator, and they do not communicate directly with each other. Power is delivered through a tether. The update frequencies for both WBDC and the torque feedback controllers are 1\si{\kilo\hertz}, and should be faster in the near future.  

In the experiments, we use our disturbance-observer (DOB) based torque feedback controller \cite{Kim:2017vl}, which accomplishes both accurate torque tracking and large phase margin. High-fidelity joint torque control is important for successful implementations of WBC, especially for this robot since it can only obtain body posture control by exploiting dynamics (the lack of the ankle actuation does not allow the robot to stand statically). 

\subsection{Body Pose Control Test}
\label{sec:exp_body_pose}

In the first experiment, we used a single task to control body orientation and CoM position. The purpose of this test was to verify that the reaction forces satisfy the contact constraints and that WBDC controls the operational space task as we intended. As mentioned in Section.~\ref{sec:wbdc_contact}, the first task needs to span the six floating base DOF (even though we know some of them are uncontrollable). We allocate small weights ($\bm{Q}_2$ in Eq.~\eqref{eq:qp_cost}) for ${\rm CoM}_x$, ${\rm CoM}_y$, and body yaw control to emphasize the importance of CoM height, body roll, and body pitch in the QP. During the experiment, a person lightly holds the robot torso and constrains the horizontal movements. Mercury controls the other aspects of its motion, such as height, roll, and pitch. To test the robot's compliance to horizontal disturbances, the experimenter moves the body left and right, and forward and backward (Fig.~\ref{fig:body_pos_ctrl}(a)).

The reaction forces presented in Fig.~\ref{fig:body_pos_ctrl}(b)(c) are commanded reaction forces since we cannot measure the reaction forces directly (the robot does not have force sensors at the feet). However, the commanded reaction forces satisfy the friction cone constraint as we can see in Fig.~\ref{fig:body_pos_ctrl}(b)(c). Fig.~\ref{fig:body_pos_ctrl}(e) presents the tracking performance of joint torque controller, and we can see the commanded and sensed torques almost overlap. The CoM position data in Fig.~\ref{fig:body_pos_ctrl}(d) shows that Mercury accurately maintains the commanded height. As we intended, the horizontal position is not restrained by the commanded position. 

\subsection{Stepping Test}
The purpose of this test is to show how the contact transition method (Section.~\ref{sec:contact_transition}) mitigates the jerk caused by a sudden change in the contact constraints while ensuring that our controller maintains the CoM height and body orientation during a stepping motion. Fig.~\ref{fig:stepping_test}(a) shows how we conducted the experiment. As we did in the body pose control experiment, a person gently holds the handles on the torso to constrain lateral movement. Stepping motion consists of three phases: double contact, transition, and swing. The duration of each phase was set to $0.055 \si{\second}$ for transitions, $0.01 \si{\second}$ for double contact, $0.36 \si{\second}$ for swings. 

Fig.~\ref{fig:stepping_test}(b) shows that the vertical reaction forces smoothly change from $0$ to $200\si{\newton}$ during transition periods (blue box). Without the described transition smoothing, the reaction force commands will suddenly jump from $0$ to $200\si{\newton}$ when contact constraints switch. The controlled position and orientation data are presented in Fig.~\ref{fig:stepping_test}(c). The $x$ and $y$ direction CoM and yaw orientation are not presented since they are uncontrolled dimensions. We use a quaternion for the body orientation description, so we show the $q_x$ and $q_y$, which are $x$ and $y$ components of a quaternion $[q_{\omega}, q_x, q_y, q_z]$, instead of roll and pitch, respectively. We can see that the body orientation stays close to the commanded orientation during the stepping motion. The experimental results show that the proposed WBDC stably and accurately control stepping motions, which is a promising foundation for the upcoming untethered walking tests. 

\section{Conclusion}
In this paper, we propose a new WBC formulation that is computational efficient, algorithmically robust, and technically sound. The proposed algorithm ensures internal and contact constraints are satisfied while conforming to a strict task hierarchy. It uses QP to compute reaction forces satisfying contact constraints and then addresses operational space tasks defined as desired accelerations. Since we utilize acceleration-based prioritized task execution instead of computing operational space dynamics, the whole projection process is concise and efficient. In our physics-based simulation tests, WBDC computes torque commands for a 28-DoF humanoid robot in $0.6\si{\milli\second}$ and the computation time is constant even when the number of tasks increases. The proposed methods were successfully implemented in our biped robot and the experiments demonstrated body posture control and smooth stepping behavior.

We are currently working on untethered walking with the passive ankle biped, which requires a robust planning algorithm and high-performance feedback controllers. By integrating the new WBDC and DOB-based torque feedback controller with our robust walking planning algorithm, we are aiming to show 3D untethered walking soon.

\section*{Acknowledgments}
The authors would like to thank the members of the Human Centered Robotics Laboratory at The University of Texas at Austin for their great help and support. This work was supported by the Office of Naval Research, ONR Grant [grant \#N000141512507] and NASA Johnson Space Center, NSF/NASA NRI Grant [grant \#NNX12AM03G].

\bibliographystyle{IEEEtran}
\bibliography{wbdc}

\end{document}